\def\year{2020}\relax
\documentclass[letterpaper]{article} 
\usepackage{aaai20}  
\usepackage{times}  
\usepackage{helvet} 
\usepackage{courier}  
\usepackage[hyphens]{url}  
\usepackage{graphicx} 
\usepackage{graphicx}  
\usepackage{times}
\usepackage{epsfig}
\usepackage{amsmath}
\usepackage{amssymb}
\usepackage{multirow}
\usepackage{algorithm}
\usepackage{algorithmic}
\usepackage{threeparttable}
\usepackage{caption}
\usepackage{subfigure}
\usepackage[switch]{lineno}
\title{SSAH: Semi-supervised Adversarial Deep Hashing with \\Self-paced Hard Sample Generation}
\author{Sheng Jin \thanks{This work was done as a research intern in Alibaba Group.}\textsuperscript{\rm 1,2},
	Shangchen Zhou\textsuperscript{\rm 3},
	Yao Liu\textsuperscript{\rm 2},
	Chao Chen\textsuperscript{\rm 2},\\
	\Large\textbf{Xiaoshuai Sun\textsuperscript{\rm 4},
	Hongxun Yao\thanks{Correspondence Author}\textsuperscript{\rm 1},
	Xian-Sheng Hua\textsuperscript{\rm 2}}\\
	\textsuperscript{\rm 1}The Harbin Institute of Technology,
	\textsuperscript{\rm 2}Alibaba DAMO Academy, Alibaba Group\\
	\textsuperscript{\rm 3}Nanyang Technological University,
	\textsuperscript{\rm 4}Xiamen University\\
	jsh.hit.doc@gmail.com, h.yao@hit.edu.cn.}
 
\begin{document}
\maketitle

\begin{abstract}
Deep hashing methods have been proved to be effective and efficient for large-scale Web media search. The success of these data-driven methods largely depends on collecting sufficient labeled data, which is usually a crucial limitation in practical cases. The current solutions to this issue utilize Generative Adversarial Network (GAN) to augment data in semi-supervised learning. However, existing GAN-based methods treat image generations and hashing learning as two isolated processes, leading to generation ineffectiveness. Besides, most works fail to exploit the semantic information in unlabeled data. In this paper, we propose a novel \textbf{S}emi-supervised \textbf{S}elf-pace \textbf{A}dversarial \textbf{H}ashing method, named \textbf{SSAH} to solve the above problems in a unified framework. The SSAH method consists of an adversarial network (A-Net) and a hashing network (H-Net). To improve the quality of generative images, first, the A-Net learns hard samples with multi-scale occlusions and multi-angle rotated deformations which compete against the learning of accurate hashing codes. Second, we design a novel self-paced hard generation policy to gradually increase the hashing difficulty of generated samples. To make use of the semantic information in unlabeled ones, we propose a semi-supervised consistent loss. The experimental results show that our method can significantly improve state-of-the-art models on both the widely-used hashing datasets and fine-grained datasets.
\end{abstract}

\section{Introduction}
In the big data era, large-scale image retrieval is widely used in many practical applications, yet it remains a challenge because of the large computational cost and high accuracy requirement. To address the efficiency and effectiveness issues, hashing methods have become a hot research topic. A great number of hashing methods are proposed to map images into a hamming space, including \textit{traditional hashing methods} \cite{andoni2006near,lin2018supervised,lin2019towards} and \textit{deep hashing methods} \cite{cao2017deep,cite:CVPR18DCH,liu2018deep,jin2018saliency}. Compared with traditional ones, deep hashing methods usually achieve better retrieval performance due to its powerful ability of feature representation and nonlinear mapping.

\begin{figure}
	\centering
	\includegraphics[width=0.95\linewidth]{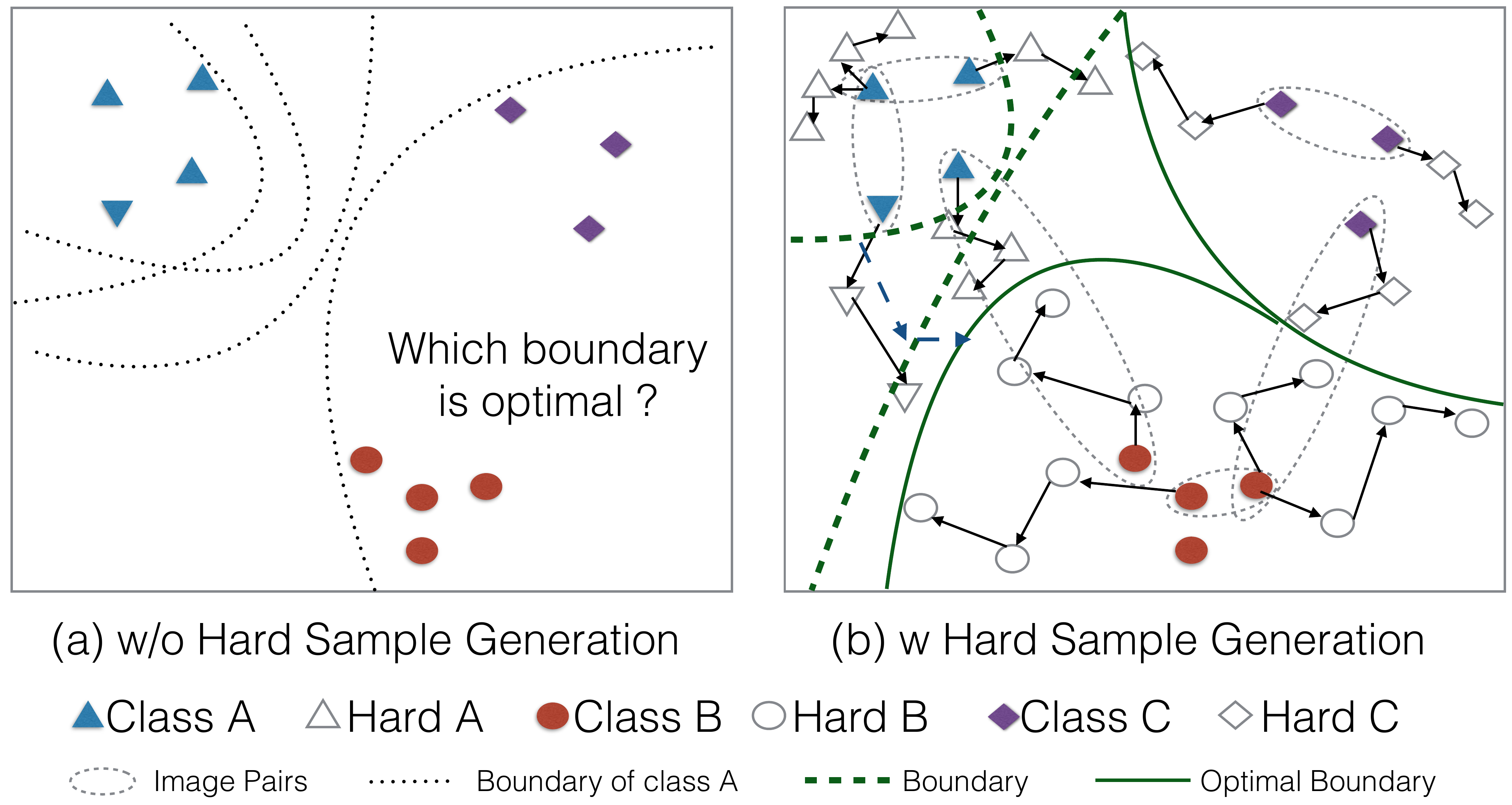}
	\caption{To obtain the optimal boundary for points with similar hashing codes, we propose a novel self-paced deep adversarial hashing to generate hard samples, as shown in (b). Intuitively, these samples can help the network learn more optimal classification boundaries.}
	\label{Motivation}
\end{figure}
Although great efforts have been devoted to deep learning-based algorithms, the label-hungry property makes it intractable in practice. Contrarily, for some retrieval tasks, unlabeled data is always enough. To make use of unlabeled data, several semi-supervised methods are proposed, include \textit{graph-based methods} like SSDH \cite{zhang2017ssdh} and BGDH \cite{yan2017semi}, and \textit{generation methods} like DSH-GANs \cite{qiu2017deep}, HashGAN \cite{cao2018hashgan} and SSGAH \cite{wang2018semi}. Graph-based works like SSDH and BGDH use graph structure to mine unlabeled data. However, constructing the graph model of large-scale data is expensive computation and time-consuming, and using batch data instead may lead to a suboptimal result. Currently, GAN \cite{goodfellow2014generative} is proved to be effective in generation tasks and then this novel technical are introduced into hashing. Existing GAN-based methods restricted by two crucial problems, \textit{i.e.}, \textbf{generation ineffectiveness} and \textbf{unlabeled-data underutilization}. 

In terms of \textbf{generation ineffectiveness}, the existing \textit{GAN-based} methods train the generation network solely based on label information. This setting leads to ineffective generations that are either too hard or easy for hashing code learning, which unable to match the dynamic training of the hashing network. In terms of \textbf{unlabeled-data underutilization}, most existing works like DSH-GANs \cite{qiu2017deep} only exploit unlabeled data to synthesize high-quality images, while the unsupervised data is not utilized when learning hashing codes. We argue that, the above two issues are not independent. In particular, the ineffecitive generation policy makes triplet-wise methods like SSGAH \cite{wang2018semi} failed to make the most of unlabeled data, since these algorithms heavily depend on hard triplets. 

In this paper, we propose a novel deep hashing method as a solid solution for \textbf{generation ineffectiveness} and \textbf{unlabeled-data underutilization} termed semi-supervised self-paced deep adversarial hashing (\textbf{SSAH}). The main idea of SSAH is depicted in Figure~\ref{Motivation}.

To tackle \textbf{generation ineffectiveness}, first, our method tries to generate proper hard images to gradually improve hashing learning. The generation network is designed to produce hard samplesL\footnote[1]{We define the samples which are difficult for current retrieval as hard samples.} with multi-scale masks and multi-angle rotations. Second, we apply the key idea of SPL\footnote[2]{The SPL theory \cite{kumar2010self} is inspired by the learning process of human, where samples are involved in learning from easy to gradually complex ones.} to our framework aiming to control the hard generations in the dynamic training procedure. To tackle \textbf{unlabeled-data underutilization}, we propose a consistent loss by encouraging consistent binary codes for the input image (both labeled and unlabeled data) and its corresponding hard generations.

Specially, \textbf{SSAH} consists of an adversarial generation network (A-Net) and a hashing network (H-Net). The loss function contains four components, including a self-paced adversarial loss, a semi-supervised consistent loss, a supervised semantic loss, and a quantization loss, which guide the training of two networks in an adversarial manner. The A-Net learns deformations and masks to generate hard images, where the quality of these generative images are evaluated by the H-Net. Then the H-Net is trained by both the input images and the generated hard samples. In the test phase, we only use the H-Net to produce hashing codes. 

The main contributions of \textbf{SSAH} are three-fold:
\begin{itemize}
	\item To generate samples properly, we propose a novel hashing framework by integrating the self-paced adversarial mechanism into hard generations and hashing codes learning. Our generation network takes both masking and deformation into account. 
	
	\item To make better use of unlabeled data, a novel consistent loss is proposed to exploit semantic information of all the data in a semi-supervised manner. 
	
	\item Experimental results on both general and fine-grained datasets demonstrate the superior performance of our method in comparison with many state-of-art methods. 
\end{itemize}
\section{Related Work}
We introduce the most related works from two aspects: \textit{Semi-supervised Hashing} and \textit{Hard Example Generation}.
\begin{figure*}
	\centering
	\includegraphics[width=0.95\linewidth]{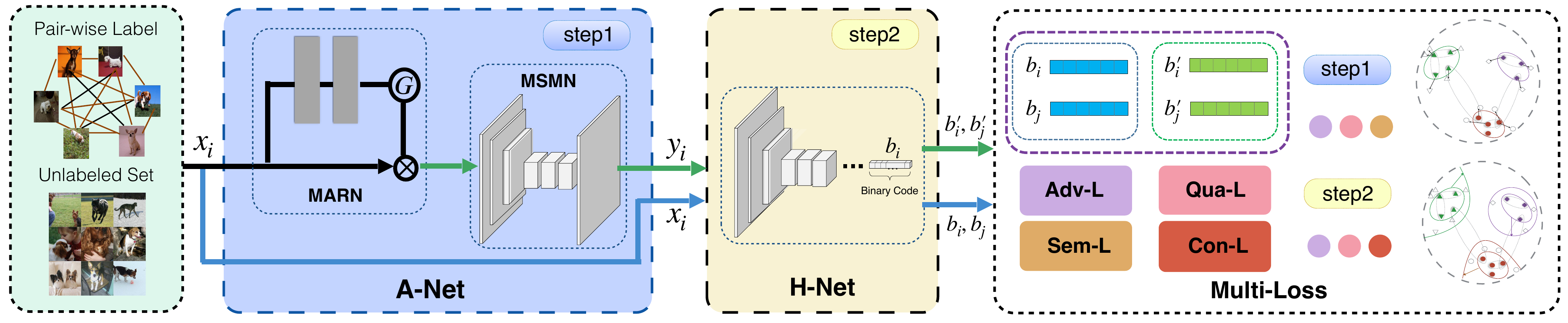}
	\caption{An illustration of self-paced adversarial hashing (\textbf{SSAH}). \textbf{SSAH} is comprised of two main components: (1) an adversarial network (A-Net) for hard sample generation, (2) a hashing network (H-Net) based on AlexNet for learning hashing codes. In the step1, the A-Net takes as input the training images and pairwise similarity to learn a hard mask. This mask is learned under the criterion that the hashing codes of masked image pairs become contrary to their pairwise label. In the step2, the H-Net take both training images and the generated hard samples as input to learn more accurate binary codes. In the training stage, hard generation and hashing codes are learned in an adversarial way.}
	\label{Fig:framework}
\end{figure*}


\noindent\textbf{Semi-supervised Hashing} Based on whether labeled data is used in the training process, hashing methods can be divided into unsupervised \cite{liu2011hashing}, semi-supervised \cite{yan2017semi}, and supervised ones \cite{xia2014supervised}. Semi-supervised hashing is effective when a small amount of labeled data and enough unlabeled data is available. SSH \cite{wang2010semi} is proposed to minimize the empirical error over labeled data and maximize the information entropy of binary codes over both labeled and unlabeled data. However, SSH is a traditional shallow method, which leads to unsatisfying performance compared with deep hashing methods like SSDH \cite{zhang2017ssdh}, BGDH \cite{yan2017semi}, which are discussed in the introduction. 

Very recently, some semi-supervised hashing methods are proposed, which use GAN \cite{goodfellow2014generative} to augment data. Deep Semantic Hashing (DSH-GANs) \cite{qiu2017deep} is the first hashing method that introduces GANs into hashing. But it can only incorporate pointwise label which is often unavailable in online image retrieval applications. Cao \textit{et al.} propose a novel conditional GANs based on pairwise supervised information, named HashGAN \cite{cao2018hashgan} to solve the insufficient sample problem. However, the sample generation is independent of hashing codes learning. Wang \textit{et al.} \cite{wang2018semi} propose SSGAH which utilizes triplet-labels and specifically designs a GANs model which can be well learned with limited supervised information. For cross-model hashing, Zhang \textit{et al.} \cite{zhang2018attention} mines the attention region in an adversarial way. However, all the mentioned GAN-based methods try to generate as much as possible images, which is not an effective and even feasible solution in most cases.


\noindent{\bfseries Hard Example Generation.} 
Hard example generation is currently used in training deep models effectively for many computer vision tasks, including object detection \cite{wang2017fast}, retrieval \cite{huang2018adversarially}, and other tasks \cite{peng2018jointly}. Zhong \textit{et al.} \cite{zhong2017random} propose a parameter-learning free method, termed Random Erasing which randomly selects a rectangle region in an image and erases its pixels with random values. However, hard example generations of Random Erasing is still isolated with network training. The generated images may not be consistent with the dynamic training status. Very recently, Wang \textit{et al.} \cite{wang2017fast} introduce the adversarial mechanism to synthesize hard samples. This method incorporates pointwise supervised information, \textit{e.g.} class labels, which is often unavailable in online image retrieval applications. Different from previous methods, we propose a novel architecture using the pairwise label to generate hard samples for learning better hashing codes. What’s more, our proposed generation networks learn hard samples in a self-paced adversarial learning manner.

\section{Semi-supervised Self-paced Adversarial Hash}
Given a dataset consist of labeled and unlabeled data. Since each labeled image in the dataset owns a unique class label, image pairs can be further labeled as similar or dissimilar $S_{ij} = 0 \text{ or } 1$.
where $S_{ij}$ denotes the pairwise label of images ($x_i$, $x_j$). Our goal is to learn more accurate hashing codes $b_i$ in a semi-supervised way. To this end, we will present \textbf{S}emi-supervised \textbf{S}elf-paced \textbf{A}dversarial \textbf{H}ashing (\textbf{SSAH}) in details. Since discrete optimization is difficult to be solved by deep networks, we firstly ignore the binary constraint and concentrate on binary-like code $\mu_i$. Then we obtain $b_i$ from $\mu_i$. Figure ~\ref{Fig:framework} illustrates an overview of our architecture, which consists of an A-Net for hard samples generation and an H-Net for binary codes learning. The generation network takes labeled and unlabeled images as inputs and produces hard masks and rotated-related parameters as the outputs. Then the H-Net learns compact binary hashing codes from both generative hard samples and training images. 
\subsection{Adversarial Network}
The A-Net is designed to generate hard images. Our method generates hard images in two main methods. The first method is to change the rotation angle and scale of the whole image. Here we propose Multi-Angle Rotation Network, termed \textbf{MARN}. The second method is to generate masks to change the value of the pixel. Here we propose Multi-Scale Mask Network, termed \textbf{MSMN}. 

\noindent \textbf{Multi-Angle Rotation Network.} Motivated by STN \cite{jaderberg2015spatial}, we propose the MARN to create multi-angle rotations on the training images. The reason is we want to preserve the location correspondence between image pixels and landmark coordinates. Otherwise, we might hurt the localization accuracy once the intermediate feature maps are disturbed. However, the Single-angle RN will often predict the largest angle. To improve the diversity of generated images, the MARN is designed to produce $n$ hard generations with size $d_0\times d_0\times (3\times n)$, where $d_0$ is the scale of images $x_0$. Each generated hard samples are required in different ranges of angles. More specially, the rotation degree of the first image is constrained within $10^{\circ}$ clockwise and anti-clockwise. The $n$-th generated image is constrained within $[10^{\circ}\times (n-1), 10^{\circ}\times n]$ clockwise and anticlockwise. For each input image $x_i$, the MARN produces $n$ hard generated images, which is denoted as ${y_i}^{(n)}$. 


\noindent \textbf{Multi-Scale Mask Network.} The object of MSMN is to produce multi-scale masks to make training images harder. In the hashing learning stage, we can obtain the convolutional features from different layers in H-Net. These features represent different spatial scales region of the original image. Correspondingly, we generate multi-scale additive and multiplicative masks for these selected features.

\begin{figure}
	\centering
	\includegraphics[width=0.95\linewidth]{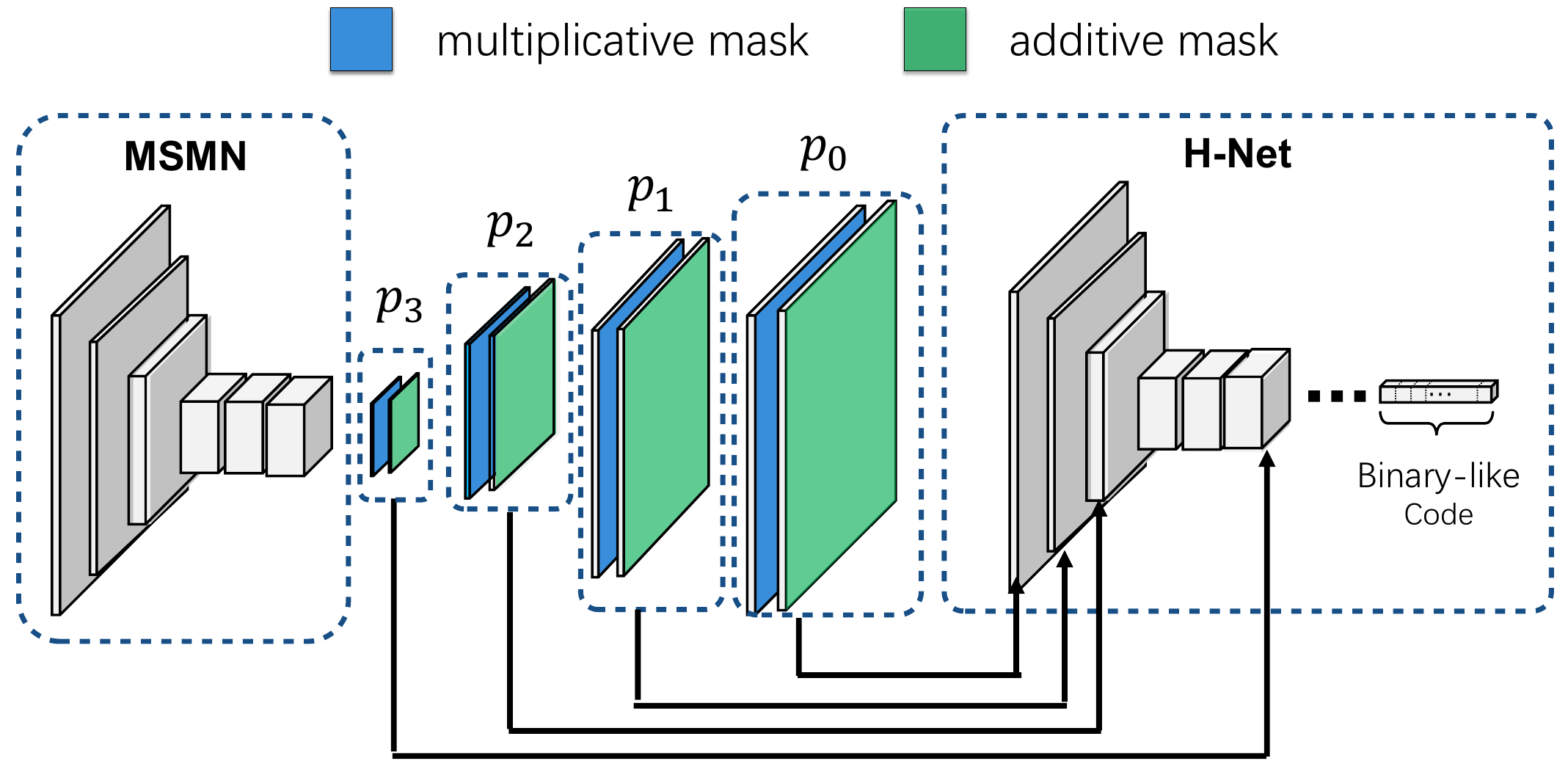}
	\caption{The framework of MSMN, where $p_i$ denotes multi-scale masks for the selected layer.}
	\label{Fig:MSMN}
\end{figure}

The framework of MSMN is shown in Figure~\ref{Fig:MSMN}. Specially, for each selected convolutional layer $m$, we extract features ${f}_m$ with size $d_m\times d_m\times c_m$, where $d_m$ is the spatial dimension and $c_m$ represents the number of channels. Given this feature, our MSMN predicts an additive hard mask $AM_m$ and a multiplicative hard mask $PM_m$.

We use the sigmoid function as the activation function for additive masks and tanh function for multiplicative masks. The value of $AM_m$ with $d_m\times d_m$ is in range of $[0, 1]$ and that of $PM_m$ is in range of $[-1, 1]$. The corresponding features of hard samples, which is denoted as ${F}_m$, is obtained by
\begin{equation}
\label{hard_feature}
F_m= (I-sigmoid(PM_m)) \cdot f_m + tanh(AM_m),
\end{equation} 
When the value of $m$ is 0, $f_m$ represents the images $y_i^{(n)}$. 
\subsection{Hashing Learning Network}
We directly adopt a pre-trained AlexNet \cite{krizhevsky2012imagenet} as the base model of the H-Net. The raw image pixel, from either the original images and the generated images, is the input of the hashing model. The learned additive masks $AM_m$ and multiplicative masks $PM_m$ are only required in the coding process of generated images $y_i$. The hard feature maps are computed according to Eq~(\ref{hard_feature}). For each generated images $y_i$, $\mu_i$ are learned:
\begin{equation}
\label{eq:trainingcodes}
\begin{aligned}
{{\mu'}_i}^{(n)} = \textit{H-Net}({y_i}^{(n)}| {AM_m, PM_m}, \forall m >0).
\end{aligned}
\end{equation}
The output layer of AlexNet is replaced by a hashing layer where the dimension is defined based on the length of the required hashing code.
\subsection{Loss Functions}
In this section, we design the adversarial loss function, including a self-paced adversarial loss and a supervised semantic loss, for the A-Net to guide the generations of hard samples. Besides, a hashing loss function for the H-Net, including the supervised semantic loss and the semi-supervised consistent loss, is proposed to learn better codes by capturing semantic information of all data. 

\noindent\textbf{Self-paced Adversarial Loss.}
Most existing GAN based hashing methods try to generate images to augment data. However, these methods can not ensure the quality of generative samples, which may obtain bad samples: (1) too difficult or too easy, (2) unable to match the dynamic training. 

To improve the \textbf{effectiveness of generations}, in this paper, we leverage the H-Net to guide the training of the A-Net by a novel definition termed \textbf{hard degree}. Firstly, we compute the similarity degree of image pairs ($x_i$, $x_j$) by using $\frac{\mu_{i}^T* \mu_{j}+k}{2k}$, where $\mu_i$ represents binary-like codes of the input images. The distance of the pairwise label and the estimated similarity degree is denoted as $d_{ij}$: 
\begin{equation}
d_{ij}= S_{i,j}-(2*S_{i,j}-1)*\frac{\mu_{i}^T* \mu_{j}+k}{2k} .
\end{equation}
$d'_{ij}$ can be obtained similarly, where $\mu'_{i}$ represents the binary-like codes of their corresponding generated images. .

Since the hard samples learned by the A-Net may increase the H-Net loss, the value of ${d'}_{ij}$ is required to be larger than that of $d_{ij}$. The \textbf{hard degree} $hard(y_i,y_j)$ of generative image pairs ($y_i$, $y_j$) is defined by using the difference between $d_ij$ and $d'_{ij}$, which can be formulated as:
\begin{equation}
\label{hard_degree}
hard(y_i, y_j)= d'_{ij}-d_{ij}=\frac{(2S_{i,j}-1)}{2k}({{\mu'}_{i}^T{\mu'}_{j}-\mu_{i}^T \mu_{j}}).
\end{equation}

We adopt a positive dynamic margin $(1-{d}_{ij})\omega$ to define the self-paced adversarial loss, which can be concluded as:
\begin{equation}
\label{hard_adv}
{l}_{adv}({y}_{i}, {y}_{j}, \omega)=\max (\omega(1-d_{ij})-hard(y_i, y_j),0).
\end{equation}
where $\omega >0$ is a fixed constant.

\textbf{Discussion.} The self-paced adversarial loss has two merits: \textit{considering inter-pair difference} and \textit{generating hard samples gradually}. In terms of \textit{inter-pair difference}, the value of ${d}_{ij}$ is large, when original image pairs $(x_i,x_j)$ is hard to distinguish. In this situation, the hard degree of $(y_i,y_j)$ is not necessarily large. By contrast, if $(x_i,x_j)$ is easy to be distinguished, the hard degree of $(y_i,y_j)$ need to be relatively large. To meet difference requirements, $(1-{d}_{ij})$ adapts the margin of hard degree. In terms of \textit{hard generations policy}, with the training of H-Net, better codes are learned and then $d_{ij}$ become small gradually. $1-{d}_{ij}$ can be used to obtain a larger margin, leading to harder generations.

Similarly, we use $x_i, y_j$ as cross-domain image pairs, where $x_i$ is the input image and $y_i$ is its corresponding hard samples. The \textbf{hard degree} of ($x_i$, $y_j$) is define as $hard(x_i, y_j)$, which can be formulated as:
\begin{equation}
hard(x_i, y_j)= \frac{(2S_{i,j}-1)}{2k}({{\mu}_{i}^T{\mu'}_{j}-\mu_{i}^T\mu_{j}}).
\end{equation}

Then self-paced adversarial loss of ($x_i$, $y_j$) is defined as ${l}_{adv}({x}_{i}, {y}_{j}, \omega/2)$, where the constant $\omega >0$ is used in Eq~(\ref{hard_adv}).
The whole self-paced adversarial loss is written as:
\begin{equation}
\label{eq:hardlossadv}
\begin{aligned}
{l}_{sp}= \sum_{i, j} \sum_{n} {l}_{adv}({y}^{(n)}_{i}, {y}^{(n)}_{j}, \omega)+{l}_{adv}({x}_{i}, {y}^{(n)}_{j}, \omega/2).
\end{aligned}
\end{equation}

\noindent \textbf{Supervised Semantic Loss.} 
Similar to other hashing methods, we adopt a pairwise semantic loss to ensure the hashing codes preserve relevant class information. The similarity degree of image pairs is computed by $\frac{{\mu}_{i}^T* \mu_{j}+k}{2k}$. Then the value of similarity degree is required to near the pairwise label $S_{i,j}$. Since the H-Net is trained by both labeled image and their corresponding hard generations, the supervised semantic loss can be written as:
\begin{equation}
\label{eq:semantichash}
\begin{aligned}
l_{sem} &=\sum_{i, j} l_{pair} (x_i,x_j) + l_{pair} (x_i,y_j) + l_{pair} (y_i,y_i)\\
&=\sum_{i, j} \frac{{\mu}_{i}^T* \mu_{j}}{2k} + \frac{{\mu'}_{i}^T* \mu_{j}}{2k} + \frac{{\mu'}_{i}^T* \mu'_{j}}{2k} + \frac{3}{2}.
\end{aligned}
\end{equation}
Besides, when training the A-Net, the supervised semantic loss is also adopted as a \textbf{regular term} to preserve class information of the generative hard samples.

\noindent \textbf{Semi-supervised Consistent Loss.}
In some related computer vision tasks like semi-supervised classification, pseudo-ensembles methods\footnote[2]{These methods develop from the cognitive ability of human. When a percept is changed slightly, a human typically still consider it to be the same object.} \cite{bachman2014learning,laine2016temporal,tarvainen2017mean} are proved to be effective. These methods encourage consistent networks output for each image with and without noise. Motivated by the success of these works, we propose a novel consistent loss to improve the \textbf{utilization of unlabeled data}. 

However, compared with existing pseudo-ensembles methods, which always adopt random and data-independent noise, our proposed method designs the A-Net to generate more proper noise for the inputs, including multi-scale masks and multi-angle rotation. Then the H-Net is required to learn consistent binary codes of the training images $x_i$ (including labeled and unlabeled images) and their corresponding hard samples $y_i$, by taking the hamming distance between the hashing codes ${\mu}_{i}$ and ${\mu'}_{i}$. The consistent loss can be formulated as:
\begin{equation}
\label{eq:conlosshash}
l_{con}= \sum_{i} {{\left \Vert \frac{k-{\mu'}_{i}^T* \mu_{i}}{2k}\right \Vert}}_{2}.
\end{equation}

\noindent \textbf{Quantization Loss. } 
Due to the ignorant of the binary constraint, a widely-used quantization loss is used to pull the value of $\mu_i$ and that of $b_i$ together, which is written as:
\begin{equation}
\label{eq:quantizationlossadv}
l_{quan}=\sum_{i} {{\left \Vert \mu'_i-b'_i\right \Vert}}_{1}+{{\left \Vert \mu_i-b_i\right \Vert}}_{1},
\end{equation}
\renewcommand{\algorithmicrequire}{ \textbf{Input:}}
\renewcommand{\algorithmicensure}{ \textbf{Output:}}
\begin{algorithm}[tb]
	\caption{Self-paced Deep Adversarial Hashing}
	\label{alg:SSAH}
	\begin{algorithmic}[1]
		\REQUIRE Training set and their corresponding class labels.
		\ENSURE H-Net function: $\textit{H-Net}(x|\Theta_{2})$ and A-Net function: $\textit{A-Net}(x|\Theta_{1})$.
		\STATE For the entire training set, construct the pairwise label matrix $S$.
		\FOR{$t=1:T$ \textit{epoch}}
		\STATE Compute $B$ accordi.ng to Eq~(\ref{eq:trainingcodes}).
		\STATE Fixing $\Theta_2$, update $\Theta_{1}$ according to Eq~(\ref{eq:adver}).
		\STATE Fixing $\Theta_1$, update $\Theta_{2}$ according to Eq~(\ref{eq:hashing}).
		\ENDFOR
		\RETURN $HN(x|\Theta_{2})$, $AN(x|\Theta_{1})$.
	\end{algorithmic}
\end{algorithm}
\subsection{Alternating Optimization}
Our network consists of two sub-networks: an adversarial network, termed A-Net, for hard image generation and a hashing network, termed H-Net, for compact hashing codes learning. As shown in Algorithm.~\ref{alg:SSAH}, we train the A-Net and the H-Net iteratively. The overall training objective of the A-Net integrates the semantic loss defined in Eq~(\ref{eq:semantichash}), the self-paced adversarial loss of three types of image pairs defined in Eq~(\ref{eq:hardlossadv}) and the quantization loss defined in Eq~(\ref{eq:quantizationlossadv}). The A-Net is trained by the following loss:
\begin{equation}
\label{eq:adver}
\min \limits_{\Theta_{1}} \quad \alpha l_{sp} + \lambda_1 l_{sem}+ \beta l_{quan}.
\end{equation}
By minimizing this term, the A-Net is trained to generate proper hard samples, leading to better hashing codes.

For the shared hashing model, the H-Net is trained by the input data, including labeled and unlabeled images, and their corresponding hard genrative samples. The supervised semantic loss, semi-supervised consistent loss, and the quantization loss are used to train the H-Net. We update its parameters according to the following overall loss:
\begin{equation}
\label{eq:hashing}
\begin{aligned}
\min \limits_{\Theta_{2}} \quad \lambda_1 l_{sem}+ \lambda_2 l_{con}+ \beta l_{quan}.
\end{aligned}
\end{equation}
By minimizing this term, the shared H-Net is trained to learn effective hashing codes.
%
\section{Experiment}
To test the performance of our proposed \textbf{SSAH} method, we conduct experiments on general hashing datasets, \textit{i.e.} CIFAR-10 and NUS-WIDE, to verify the effectiveness of our method. Then we conduct experiments on two fine-grained datasets, \textit{i.e.} CUB Bird and Stanford Dogs-120, to prove that our method is still robust and effective for more complex fine-grained retrieval tasks. We also conduct some analytical experiments to further verify our method.
\begin{table}[tp]
	\small
	\centering
	\caption{The mAP scores for different number of bits on CIFAR-10 and NUSWIDE datasets.}
	\label{tab1:}
	\resizebox{0.95\linewidth}{!}{
	\begin{tabular}{r|ccc|ccc}
		\hline
		\multirow{2}{*}{Dataset}&
		\multicolumn{3}{c|}{CIFAR-10}&\multicolumn{3}{c}{NUSWIDE}\cr\cline{2-7}
		&12 bits&24 bits&48 bits&12 bits&24 bits&48 bits\cr
		\hline
		ITQ-CCA&0.435&0.435&0.435&0.526&0.575&0.594\cr
		KSH&0.556&0.572&0.588&0.618&0.651&0.682\cr
		SDH&0.558&0.596&0.614&0.645&0.688&0.711\cr
		CNNH&0.439&0.476&0.489&0.611&0.618&0.608\cr
		HashGAN&0.655&0.709&0.727&0.708&0.722&0.730\cr	
		SSDH&0.801&0.813&0.814&0.773&0.779&0.778\cr	
		BGDH&0.805&0.824&0.833&0.803&0.818&0.828\cr	
		SSGAH&0.819&0.837&0.855&0.835&0.847&0.865\cr			
		\hline		
		\textbf{SSAH}&{\bf 0.862}&{\bf 0.878}&{\bf 0.886}&{\bf 0.872}&{\bf 0.884}&{\bf 0.898} \cr
		\hline
	\end{tabular}}
\end{table}

\subsection{Datasets}
We conduct our experiments on two general datasets, namely CIFAR-10 and NUSWIDE. CIFAR-10 \cite{krizhevsky2009learning} is a small image dataset including 60k 32 $\times$ 32 images in 10 classes. Each image belongs to one class (6000 images per class). NUS-WIDE \cite{chua2009nus} contains nearly 270k images with 81 semantic concepts. For NUS-WIDE, we follow \cite{liu2011hashing} to use the images associated with the 21 most frequent concepts, where each of these concepts associated with at least 5,000 images. Following \cite{liu2011hashing,wang2018semi}, we randomly sample 100 images per class as the test set, and the others are as a database. In the training process, we randomly sample 500 images per class from the database as labeled data, and the others are as unlabeled data.

%
We further verify our experiments on two widely-used fine-grained datasets, namely Stanford Dogs-120 and CUB Bird. {\bfseries Stanford Dogs-120} \cite{nilsback2006visual} dataset consists of 20,580 images in 120 mutually classes. Each class contains about 150 images. {\bfseries CUB Bird} \cite{WahCUB_200_2011} includes 11,788 images in mutually 200 classes. We directly use test set defined in these datasets. The train set is used as a database. In the training process, we randomly sample 50\% images per class s from the database as labeled data, and the others are as unlabeled data. 

\subsection{Comparative Methods and Evaluation Metrics}
For the general datasets, including CIFAR-10 and NUSWIDE dataset, we compare our method (\textbf{SSAH}) with three supervised deep hashing methods: CNNH \cite{xia2014supervised}, HashGAN \cite{cao2018hashgan}, three semi-supervised deep hashing methods: SSDH \cite{zhang2017ssdh}, BGDH \cite{yan2017semi}, SSGAH \cite{wang2018semi} and three shallow methods: ITQ-CCA \cite{gong2013iterative}, KSH \cite{liu2012supervised}, SDH \cite{shen2015supervised}. For the fine-grained datasets, our method is further compared with DSaH \cite{jin2018saliency}, which is the first hashing method designed for fine-grained retrieval. 

For a fair comparison between traditional and deep hashing methods, we conduct these methods on features extracted from the fc7 layer of AlexNet which is pre-trained on ImageNet. For deep hashing methods, we use as input the original images, and adopt AlexNet \cite{krizhevsky2012imagenet} as the backbone architecture.

\noindent \textbf{Evaluation Metric.} We use Mean Average Precision (\textbf{mAP}) for quantitative evaluation. The Precision@top-N curves and Precision-Recall curves are shown in supple materials.
\subsection{Implementation Details}
\noindent \textbf{Network Design.} As shown in Figure~\ref{Fig:framework}, our model consists of an A-Net including MSMN, MARN, and a H-Net. For MSMN, we adopt a lightweight version of generator in GANimation \cite{pumarola2018ganimation}. This network contains a stride-2 convolution, three residual blocks \cite{he2016deep}, and a 1/2-strided convolution. Similar to Johnson \textit{et al.} \cite{johnson2016perceptual}, we use instance normalization \cite{ulyanov2016instance}. The figurations of MSMN are shown in the supplematerials. MARN is built upon the STN \cite{jaderberg2015spatial}. Different from STN, our transformer network produces $n$ sets of affine parameters. We adopt AlexNet \cite{krizhevsky2012imagenet} as the encoder of H-Net, fine-tune all layers but the last one are copied from the pre-trained AlexNet.

\noindent \textbf{Trainind Details.} Our \textbf{SSAH} is implemented on PyTorch and the deep model is trained by batch gradient descent. In the training stage, images are regarded as input in the form of batch and every two images in the same batch construct an image pair. Practically, we train A-Net before H-Net. If we first train H-Net, A-Net might output semantic-irrelevant hard generated images, which would be a bad sample and guide the training of hashing model in the wrong direction.

\noindent {\bfseries Network Parameters.} The value of hyper-parameter $\lambda_1$ is 1.0, $\lambda_2$ is 0.5, $\alpha$ is 0.5 and $\beta$ is 0.1. We use the mini-batch stochastic gradient descent with 0.9 momentum. We set the value of the margin parameters $\omega$ as $0.1$, which increases $0.02$ every 5 epochs. The mini-batch size of images is fixed as 32 and the weight decay parameter as 0.0005. The value of the number of the rotated hard samples $n$ is 3.
\subsection{Quantitative Results}
{\bfseries Performance on general hashing datasets.} The Mean Average Precision (mAP,\%) results of different methods for different numbers of bits on NUSWIDE and CIFAR-10 dataset are shown in Table~\ref{tab1:}. Experimental results show that \textbf{SSAH} outperforms state-of-the-art SSGAH \cite{wang2018semi} by $3.75\%$, $3.55\%$ on CIFAR10, and NUSWIDE, respectively. According to the experimental results, \textbf{SSAH} can be seen to be more effective for traditional hashing task.

\noindent {\bfseries Performance on fine-grained hashing datasets.} The fine-grained retrieval task requires methods describing fine-grained objects that share similar overall appearance but have a subtle difference. To meet this requirement, there will be greater demand for collecting and annotating data, where professional knowledge is required in some cases. Since it is more difficult to generate fine-grained objects, GAN-based hashing methods are also not effective due to the scarcity of supervised data. However, the experimental results show that our method is still robust to this task.

The mAP results of different methods on fine-grained datasets are shown in Table~\ref{tab2:}. The proposed \textbf{SSAH} method substantially outperforms all the comparison methods. Compared with existing best retrieval performance (\underline{DSaH (Alexnet)}), \textbf{SSAH} achieves absolute increases of $4.55\%$, $6.24\%$ on CUB Brid datasets and on Stanford dog datasets, respectively. Compared with the mAP results on traditional hashing task, our method is proved to achieve a significant improvement in fine-grained retrieval. What's more, we only use the H-Net to produce binary codes in the test phase and the DSaH method need to highlight the salient regions before the encoding process. Thus, our method is also more efficient in time. 
\begin{table}[htb]
	\small
	\centering
	\caption{The mAP scores for different number of bits on Stanford Dogs-120 and CUB Bird datasets.}
	\label{tab2:}
	\resizebox{0.95\linewidth}{!}{
	\begin{tabular}{r|ccc|ccc}
		\hline
		\multirow{2}{*}{Dataset}&
		\multicolumn{3}{c|}{Stanford Dogs-120}&		
		\multicolumn{3}{c}{CUB Bird}\cr\cline{2-7}
		&12 bits&24 bits&48 bits&12 bits&24 bits&48 bits\cr
		\hline
		HashGAN&0.029&0.172&0.283&0.020&0.0542&0.123\cr
		SSGAH&0.127&0.233&0.329&0.073&0.1321&0.247\cr
		DSaH&0.244&0.287&0.408&0.091&0.2087&0.285\cr
		\textbf{SSAH}&{\bf 0.273}&{\bf 0.343} &{\bf 0.478}&{\bf 0.141}&{\bf 0.265}&{\bf 0.359}\cr
		\hline
	\end{tabular}}
\end{table}
\begin{table}[tp]
	\small
	\centering
	\caption{The mAP scores under retrieval of unseen classes on CIFAR-10 and NUSWIDE datasets.}
	\label{tab-unseen:}
	\resizebox{0.95\linewidth}{!}{
		\begin{tabular}{r|ccc|ccc}
			\hline
			\multirow{2}{*}{Dataset}&
			\multicolumn{3}{c|}{CIFAR-10}&\multicolumn{3}{c}{NUSWIDE}\cr\cline{2-7}
			&12 bits&24 bits&48 bits&12 bits&24 bits&48 bits\cr
			\hline
			ITQ-CCA&0.157&0.165&0.201&0.488&0.493&0.503\cr
			SDH&0.185&0.193&0.213&0.471&0.490&0.507\cr
			CNNH&0.210&0.225&0.231&0.445&0.463&0.477\cr
			DRSCH&0.219&0.223&0.251&0.457&0.464&0.360\cr
			NINH&0.241&0.249&0.272&0.484&0.483&0.487\cr
			SSDH&0.285&0.291&0.325&0.510&0.533&0.551\cr	
			SSGAH&0.309&0.323&0.339&0.539&0.553&0.579\cr			
			\hline		
			\textbf{SSAH}&{\bf 0.338}&{\bf 0.370}&{\bf 0.379}&{\bf 0.569}&{\bf 0.571}&{\bf 0.596} \cr
			\hline
	\end{tabular}}
\end{table}

\noindent\textbf{Performance of Unseen Classes.} To further evaluate our SSAH approach, we adopt the evaluation protocol from \cite{sablayrolles2017should}. In the training process, 75\% of classes (termed set 75) are known, and the remaining 25\% classes (termed set 25) are used to for evaluation. The set 75 and set 25 are further divided into the training set and test set. Data amount in train and test set are the same. Following settings in \cite{zhang2017ssdh}, we use train75 as the training set and test25 as the test set. For general hashing retrieval, the set75 of CIFAR-10 and NUS- WIDE consist of 7 classes and 15 classes respectively, results are calculated by the average of 5 different random splits, mAP scores are calculated based on all returned images. 

The mAP scores under the retrieval of unseen classes are shown in Table~\ref{tab-unseen:}. Our SSAH method achieves the best result when retrieving unseen classes, which means that our method achieves better generalization performance to unseen class. The experimental results on fine-grained datasets are shown in supple materials.
\subsection{Ablation Study}
To further verify our method, we conduct some analysis experiments including: (1) the effectiveness of hard samples generation, (2) the analysis of each loss component, (3) the effectiveness of self-paced generation policy. 

\noindent{\bfseries Component Analysis of the Network.} We compare our MARN and MSMN with random image rotation/random mask generation strategy in training stage using the AlexNet architecture. (1) Random Image Rotation: For each image, we obtain three rotated images, where each image are rotated in the specified angle range. (2) Random Mask Generation: The values of multiplicative masks are in range of $[0, 1]$ and that of additive masks are in range of $[-1, 1]$. The $90\%$ values of multiplicative and additive masks are required in the range of $[-0.1, 0.1]$. 

We report our results for using MARN and MSMN in Table~\ref{networkanalysis}. For the AlexNet architecture, the mAP of \underline{our implemented baseline} is $75.1\%\sim79.2\%$ and $76.2\%\sim80.1\%$ on CIFAR-10 and NUS-WIDE datasets. Based on this setting, joint learning with our MARN model improves \underline{baseline} by $4.5\%$ and $4.4\%$, respectively on these datasets. Joint learning with the MSMN model improves \underline{baseline} by $7.0\%$ and $6.9\%$. As both methods are complementary to each other, combining MARN and MSMN into our model gives another boost to $86.2\%\sim88.6\%$ and $87.2\%\sim89.8\%$ on CIFAR-10 and NUS-WIDE datasets, respectively.

\begin{table}[htb]
	\small
	\centering
	\caption{The mAP scores of SSAH using different network components (MARN and MSMN). }
	\label{networkanalysis}
	\resizebox{0.9\linewidth}{!}{
	\begin{tabular}{r|cc|cc}
		\hline
		\multirow{2}{*}{Methods}&
		\multicolumn{2}{c|}{CIFAR-10}&
		\multicolumn{2}{c}{NUSWIDE}\cr\cline{2-5}
		&12 bits&48 bits&12 bits&48 bits\cr
		\hline
		\underline{baseline}&\underline{0.751}&\underline{0.792}&\underline{0.762}&\underline{0.801}\cr
		random rotate&0.788&0.810&0.797&0.828\cr
		+MARN&0.801&0.831&0.810&0.842\cr
		\hline
		random mask&0.792&0.813&0.806&0.811\cr		
		+MSMN ($+$)&0.811&0.838&0819&0.832\cr
		+MSMN ($\times$)&0.820&0.847&0.831&0.849\cr			
		+MSMN&0.830&0.853&0.840&0.861\cr
		\hline
		random rotate+mask&0.796&0.816&0.810&0.833\cr
		\textbf{Ours(full)}&\textbf{0.862}&\textbf{0.886}&\textbf{0.872}&\textbf{0.898}\cr
		\hline
	\end{tabular}}
\end{table}

\begin{figure}[t]
	\centering
	\includegraphics[width=0.95\linewidth]{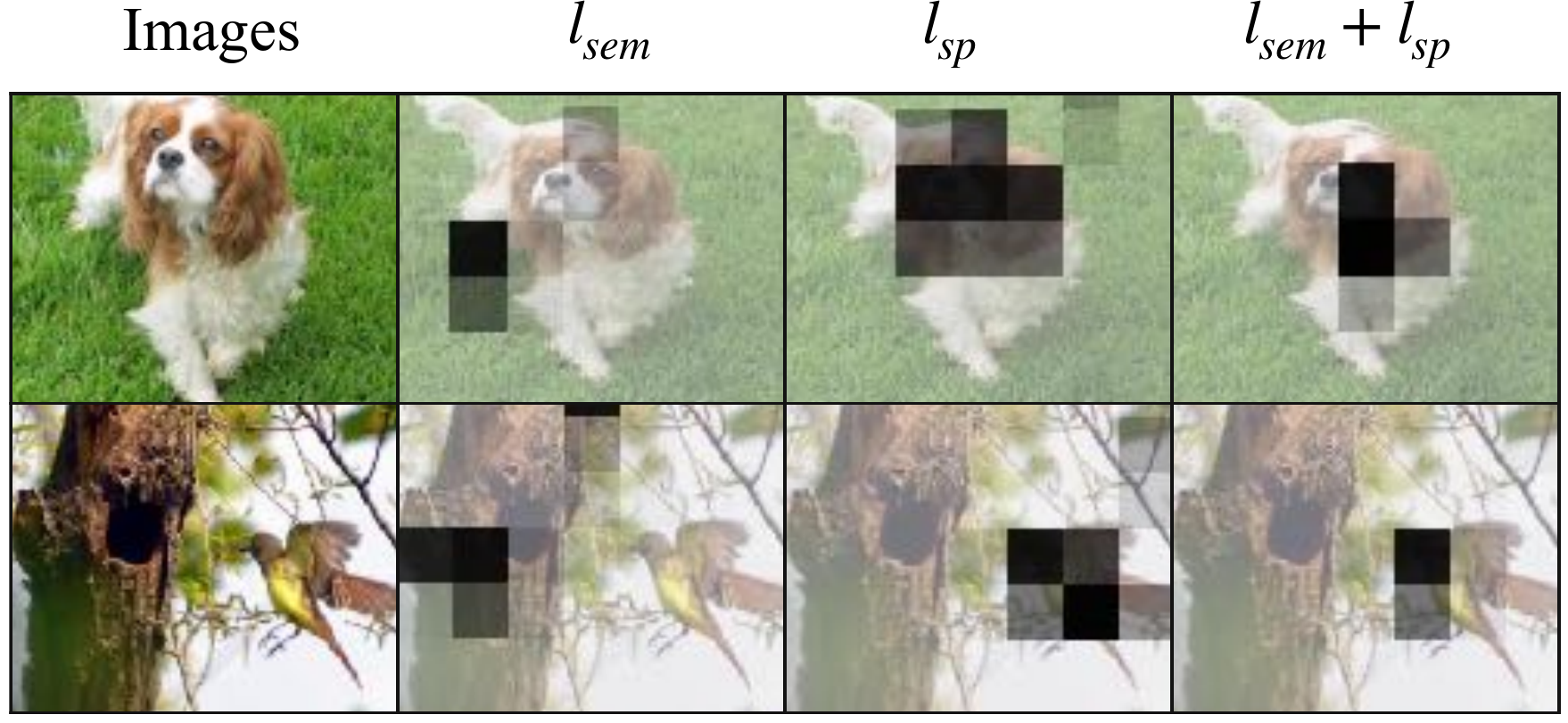}
	\caption{Typical Examples of the learned mask using different loss components on fine-grained dataset.}
	\label{Fig:loss-mask}
\end{figure}

\noindent {\bfseries Component Analysis of the Loss Functions. }Our loss function consists of three major components: $l_{sem}$, $l_{con}$ and $l_{sp}$. $l_{sp}$ includes that of hard and cross-domain image pairs, which are denoted as $l_{sph}$ and $l_{spc}$. To evaluate the contribution of each loss, we study the effect of different loss combinations on retrieval performance. From Table~\ref{lossanalysis}, when we use $l_{sem}$, $l_{con}$ and $l_{quan}$ to train H-Net, and $l_{sp}$ and $l_{sem}$ to train A-Net, the retrieval performance is best. For the H-Net, the self-paced adversarial loss $l_{sp}$ may destroy the training procedure of accurate binary codes. Combined with $l_{con}$, our method further improves about $2\%$, which shows H-Net capture the semantic information of unlabeled data. We also show some typical visualization results of hard masks using different loss components of the A-Net in Figure~\ref{Fig:loss-mask}. 

As shown in Figure~\ref{Fig:loss-mask}, for the A-Net, if we only use the semantic loss, the generated mask would avoid the object. if we only use the self-paced adversarial loss, the generated mask occludes the object in some cases. However, using the combination of $l_{adv}$ and $l_{sem}$ can obtain a proper mask. The object is partially occluded and it still can be recognized.
\begin{table}[htb]
	\small
	\centering
	\caption{The mAP scores of SSAH on CIFAR-10 dataset using different combinations of loss functions.}
	\label{lossanalysis}
	\resizebox{0.95\linewidth}{!}{	
	\begin{tabular}{r|r|ccc}
		\hline
		\multirow{2}{*}{A-Net}&\multirow{2}{*}{H-Net}&
		\multicolumn{3}{c}{CIFAR-10}\cr\cline{3-5}
		&&12 bits&24 bits&48 bits\cr
		\hline
		\underline{$l_{sem}$}&\multirow{2}{*}{$l_{sem}+l_{con}$}&0.809&0.823&0.85\cr
		$l_{sph}$&&0.821&0.843&0.862\cr
		\textbf{$l_{spc}$}&&0.836&0.850&0.870\cr
		\textbf{$l_{sp}$}&&0.848&0.862&0.872\cr
		\hline
		\multirow{3}{*}{$l_{sem}+l_{sp}$}
        &$l_{sem}$&0.841&0.855&0.867\cr
        &$l_{sem}+l_{con}$&{\bf 0.862}&{\bf 0.878}&{\bf 0.886}\cr
        &$l_{sem}+l_{con}+l_{sp}$&0.843&0.852&0.8612\cr
        \hline		
	\end{tabular}}
\end{table} 

\noindent\textbf{The effectiveness of Self-paced Hard Generation Policy. }
In this section, we evaluate SSAH on the impact of generation policy by self-paced vs. fixed-paced loss. To define the fixed-paced loss, the dynamic margin $(1-{d}_{ij})\omega$ in Eq~(\ref{hard_adv}) is replace by a fixed parameter $\omega$. The fixed-paced loss is formulated as ${l}_{fixed}({y}_{i}, {y}_{j}, \omega)=\max (\omega-hard(y_i,y_j),0)$.

As shown in Table ~\ref{selfpaceanalysis}, compare with fixed-paced loss, the self-paced adversarial loss is more effective and also robust to the margin parameter $\omega$. A possible reason is that the fixed-paced loss can not match the dynamic training procedure, and balance hard pairs and simple ones.
\begin{table}[htb]
	\small
	\centering
	\caption{The mAP scores of SSAH on CIFAR-10 using self-paced loss and fixed-paced loss with different margin $\omega$.}
	\label{selfpaceanalysis}
	\resizebox{0.95\linewidth}{!}{
	\begin{tabular}{r|cc|cc}
		\hline
		\multirow{2}{*}{Margin Parameters $\omega$}&
		\multicolumn{2}{c|}{Self-paced Loss}&
        \multicolumn{2}{c}{Fixed-paced Loss}
		\cr\cline{2-5}
		&12 bits&48 bits&12 bits&48 bits\cr
		\hline
		0.01&0.790&0.803&0.791&0.815\cr
		0.05&0.843&0.855&0.819&0.842\cr
		0.3&0.855&0.873&0.838&0.852\cr
		0.5&0.846&0.861&0.789&0.813\cr
		1.0&0.849&0.852&0.775&0.801\cr			
		\textbf{Ours(0.1)}&\textbf{0.862}&\textbf{0.886}&\textbf{0.839}&\textbf{0.861}\cr
		\hline
	\end{tabular}}
\end{table}
\section{Conclusion}
 To solve the data insufficiency problem, we propose a \textbf{S}emi-supervised \textbf{S}elf-paced \textbf{A}dversarial \textbf{H}ashing (\textbf{SSAH}) method, consisting of an adversarial network (A-Net) and a hashing network (H-Net). To exploit the semantic information in images, and their corresponding hard generative images, we adopt a supervised semantic loss and a novel semi-supervised consistent loss to train the H-Net. Then the H-Net is used to guide the training of A-Net by a novel self-paced adversarial loss to produce multi-scale masks and some sets of deformation parameters. The A-Net and the H-Net are trained iteratively in an adversarial way. Extensive experimental results demonstrate the effectiveness of \textbf{SSAH}. 
 
\noindent \textbf{Acknowledgements} This work was supported by the National Natural Science Foundation of China under Project No. 61772158, 61702136, and U1711265.
{\fontsize{9.5}{10.5}
 \bibliographystyle{aaai}
 \bibliography{ssahref}
}

\end{document}